\documentclass{article}

    \PassOptionsToPackage{numbers, compress}{natbib}

\usepackage[preprint]{neurips_2024}

\usepackage[utf8]{inputenc} 
\usepackage[T1]{fontenc}    
\usepackage{hyperref}       
\usepackage{url}            
\usepackage{booktabs}       
\usepackage{amsfonts}       
\usepackage{nicefrac}       
\usepackage{microtype}      
\usepackage{xcolor}         
\usepackage{graphicx}
\usepackage{amsmath}
\usepackage{wrapfig}
\usepackage{footnote}
\usepackage{tablefootnote}

\makesavenoteenv{table}
\makesavenoteenv{tabular}

\definecolor{lightred}{rgb}{1.0,0.6,0.6}

\definecolor{lightpurple}{rgb}{0.8,0.0,0.8}

\newif\ifcomments
\commentstrue 

\title{VITON-DiT: Learning In-the-Wild Video Try-On from Human Dance Videos via Diffusion Transformers }

\author{
Jun Zheng \\
Sun Yat-Sen University \\
\texttt{zhengj98@mail2.sysu.edu.cn} \\
\And 
Fuwei Zhao \\
ByteDance China \\
\texttt{zhaofuwei.777@bytedance.com}\\
\And
Youjiang Xu \\
ByteDance China \\
\texttt{xuyoujiang.01@bytedance.com}\\
\And 
Xin Dong \\
ByteDance China \\
\texttt{dongxin.1016@bytedance.com}\\
\And 
Xiaodan Liang\thanks{Corresponding author} \\
Sun Yat-Sen University \\
\texttt{xdliang328@mail.sysu.edu.cn} \\
}

\begin{document}

\maketitle

\begin{abstract}
  Video try-on stands as a promising area for its tremendous real-world potential.
  Prior works are limited to transferring product clothing images onto person videos with simple poses and backgrounds, while underperforming on casually captured videos. 
  Recently, Sora revealed the scalability of Diffusion Transformer (DiT) in generating lifelike videos featuring real-world scenarios. 
  Inspired by this, we explore and propose the first DiT-based video try-on framework for practical in-the-wild applications, named VITON-DiT. Specifically, VITON-DiT consists of a garment extractor, a Spatial-Temporal denoising  DiT, and an identity preservation ControlNet. 
  To faithfully recover the clothing details, the extracted garment features are fused with the self-attention outputs of the denoising DiT and the ControlNet. We also introduce novel random selection strategies during training and an Interpolated Auto-Regressive (IAR) technique at inference to facilitate long video generation.
  Unlike existing attempts that require the laborious and restrictive construction of a paired training dataset, severely limiting their scalability, VITON-DiT alleviates this by relying solely on unpaired human dance videos and a carefully designed multi-stage training strategy. Furthermore, we curate a challenging benchmark dataset to evaluate the performance of casual video try-on.
  Extensive experiments demonstrate the superiority of VITON-DiT in generating spatio-temporal consistent try-on results for in-the-wild videos with complicated human poses. Project page: \url{https://zhengjun-ai.github.io/viton-dit-page/}.

\end{abstract}

\section{Introduction}
\label{introduction}

Video virtual try-on systems~\cite{fwgan, mv-ton, clothformer, xu2024tunnel} aim to dress a target person in video with desired clothing while maintaining their motion and identity. It offers tremendous potential for practical uses such as e-commerce and entertainment. While video representation is more compelling, it is also more challenging. Therefore, the majority of existing work has focused on image-based try-on~\cite{XintongHan2018VITONAI,SeungHwanChoi2021VITONHDHV,liuLQWTcvpr16DeepFashion,YuyingGe2021ParserFreeVT,he2022fs_vton,NEURIPS2021_151de84c, xie2023gpvton, Xie2021TowardsSU}. The earlier approaches typically build on Generative Adversarial Networks (GANs)~\cite{NEURIPS2021_151de84c,he2022fs_vton,SeungHwanChoi2021VITONHDHV,xie2023gpvton, Xie2021TowardsSU}, containing a warping module and a try-on generator. The warping module deforms clothing to align with the human body, and then the warped garment is fused with the person image through the try-on generator. However, with the recent advent of UNet-based Latent Diffusion Models (LDMs)~\cite{ldm,controlnet,mou2023t2i,yang2022paint}, researchers' attention has gradually shifted to these emerging generative models for more groundbreaking results. A diffusion-based try-on network does not explicitly separate the warping and blending operations, but instead unifies them into a single process of cross-attention implicitly. By leveraging text-to-image pre-trained weights, these diffusion approaches demonstrate superior fidelity compared to the GAN-based counterparts.



\begin{figure}
    \centering
    \includegraphics[width=1\linewidth]{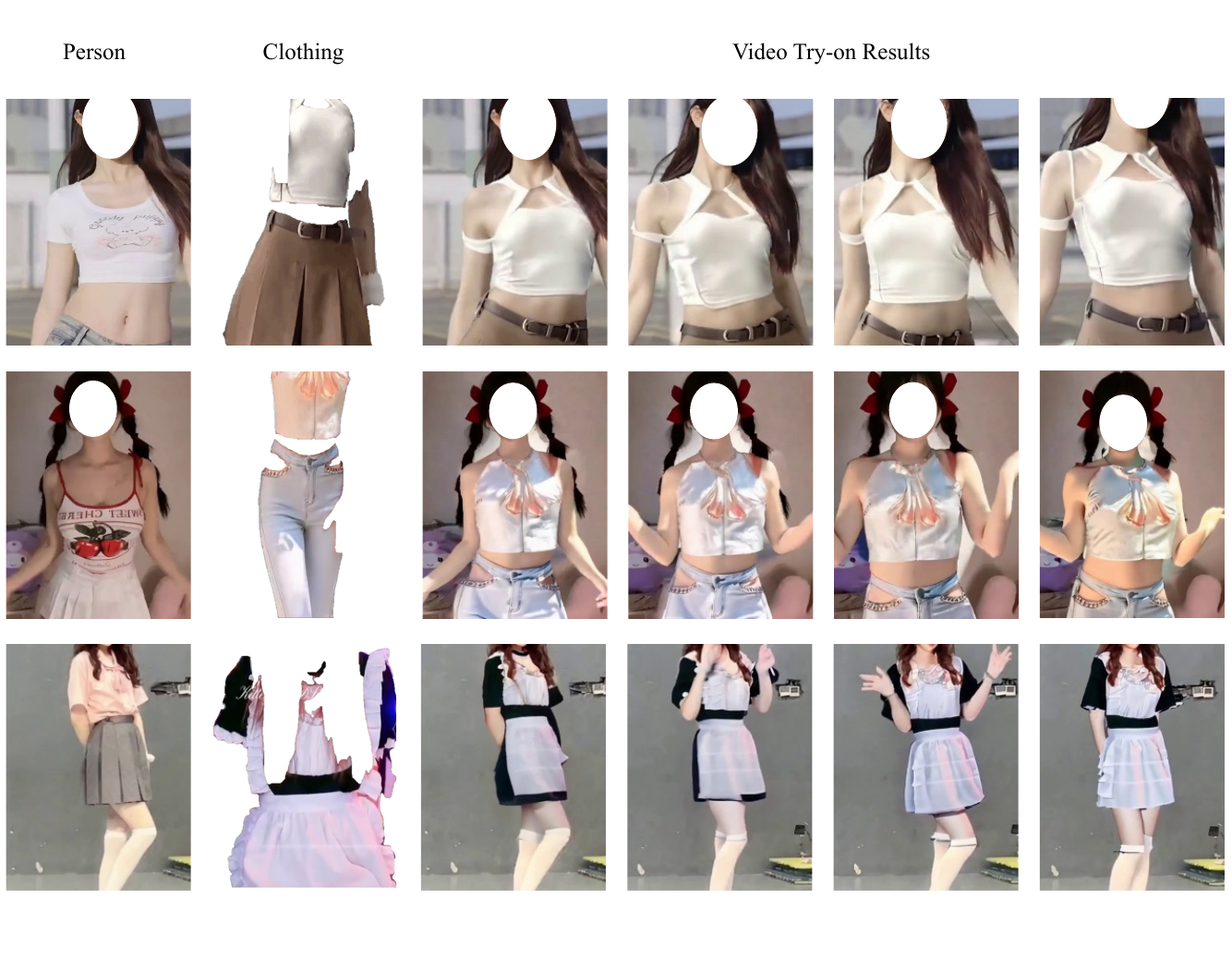}
    \vspace{-6mm}
    \caption{Video try-on results of the proposed VITON-DiT. Our model is capable of generalization across diverse types of clothing, even those non-product garments that may have flaws. It can also deal with complex body movements such as dancing against real world backgrounds.}
    \label{fig:teaser}
\end{figure}

Despite promising image generation quality, UNet-based LDMs appear to fall short when handling video scenes. Whereas the newfound Transformer-based LDMs (or Diffusion Transformer, DiT) exhibit remarkable capability and scalability in generating high-fidelity real-world images/videos, such as Stable Diffusion 3~\cite{stablediffusion3} and Sora~\cite{sora}. Inspired by Sora, we propose VITON-DiT, the first DiT-based video try-on network tailored to tackle in-the-wild scenarios. Fig.~\ref{fig:teaser} shows samples generated by our model\footnote{Please refer to the supplemented video for more results.}. Specifically, VITON-DIT contains three main components: a spatio-temporal denoising DiT for video latent generation, a garment extractor to maintain clothing details faithfully, and an ID ControlNet to preserve the person's pose and identity. These three modules are connected by an innovative \emph{attention fusion} mechanism. This mechanism combines extracted garment features with person denoising features via an addictive attention process,  enabling seamless integration of clothing characteristics into the video generation process. Furthermore, we devise a random selection strategy during training and an Interpolated Auto-Regressive (IAR) sampling technique at inference to facilitate long video generation up to tens of seconds.


Existing video try-on methods rely heavily on paired datasets~\cite{fwgan}, i.e., a product clothing image coupled with a video of a person wearing that clothing. Constructing such datasets is labor-intensive, limiting scalability and generalization ability. In contrast, our VITON-DiT only requires an unpaired collection of human dance videos, facilitated by a multi-stage training procedure. We also show that with the expansion of data volume, the model's performance scales up proportionately, highlighting its robust data scalability. Moreover, we propose a new video try-on benchmark capturing real-world complexities to comprehensively evaluate model performance.


Extensive experiments showcase VITON-DiT's superiority in generating realistic video try-on results, even for unconstrained scenes involving complex body motions and backgrounds. Our model takes the first step towards video virtual try-on for authentic real-world uses. Our main contributions include four aspects:
\begin{itemize}
    \item We present the first DiT-based video try-on network, VITON-DiT, featuring consistent spatio-temporal generation on casually captured human videos.
    \item We propose an attention fusion algorithm to bridge the three modules of VITON-DiT, enabling precise recovery of clothing details in videos.
    \item We devise a novel random selection strategy and an interpolated auto-regressive technique to foster long video generation.
    \item Our model is scalable to unpaired human video data, and we provide a new challenging video try-on benchmark to facilitate related research fields.
\end{itemize}

\section{Related Work}
\label{related_work}
\textbf{Video Virtual Try-on.} Existing work on video virtual try-on can be classified as GAN-based~\cite{fwgan,mv-ton,shineon, clothformer} and diffusion-based methods~\cite{xu2024tunnel}. The former relies on garment warping by optical flow~\cite{dosovitskiy2015flownet} and utilizes a GAN generator to fuse the warped clothing with the reference person. FW-GAN~\cite{fwgan} predicts optical flow to warp preceding frames during the video try-on process, thereby ensuring the generation of temporally coherent video sequences. ClothFormer~\cite{clothformer}  presents a dual-stream transformer architecture to efficiently integrate garment and person features, facilitating more accurate and realistic video try-on results. Despite reasonable performance, GAN-based methods struggle to garment-person misalignment in case of inaccurate warping flow estimation. And the overall generation quality is inferior to diffusion-based models that are with large-scale pre-trained weights. The recent Tunnel Try-on~\cite{xu2024tunnel} proposes to use a UNet-based diffusion model for video try-on. It can handle camera movements and faithfully preserve the clothing textures. However, their provided demo videos are only for product images and span just a few seconds. Instead, our VITON-DiT can be applied to diverse types of clothing even with missing parts, and can generate tens of seconds of try-on sequences in in-the-wild scenarios with high-quality spatiotemporal consistency.

\textbf{Diffusion Models for Video Generation}
The success of text-to-image (T2I) diffusion models leads to emerging studies on text-to-video synthesis. Such studies~\cite{2023i2vgenxl,2023videocomposer,guo2023animatediff,guo2023sparsectrl} typically insert additional temporal dimensions or layers into the pre-trained T2I models.  For example, Video LDM~\cite{vldm} proposes an innovative two-stage training process, initially focusing on static images, followed by a temporal layers specifically trained on video datasets. Similarly, AnimateDiff~\cite{guo2023animatediff} introduces a versatile plug-and-paly motion module for existing T2I models, mitigating extensive model-specific adjustments. However, these approaches still face challenges in long video generation due to the restricted capacity and scalability of the UNet architecture, whereas DiT-based T2V models~\cite{sora,opensora,opensoraplan} are capable of directly generating videos with tens of seconds. The recent phenomenal Sora~\cite{sora} adopts DiT as backbone and can generate realistic 1-minute videos from texts. However, Sora is close-sourced currently and we resort to one of its alternatives named Open-Sora~\cite{opensora}. Our VITON-DiT is built upon OpenSora and served as the first DiT-based model for video virtual try-on.

\section{Method}
\label{method}

\begin{figure*}[h!]
    \centering
    \includegraphics[width=1\linewidth]{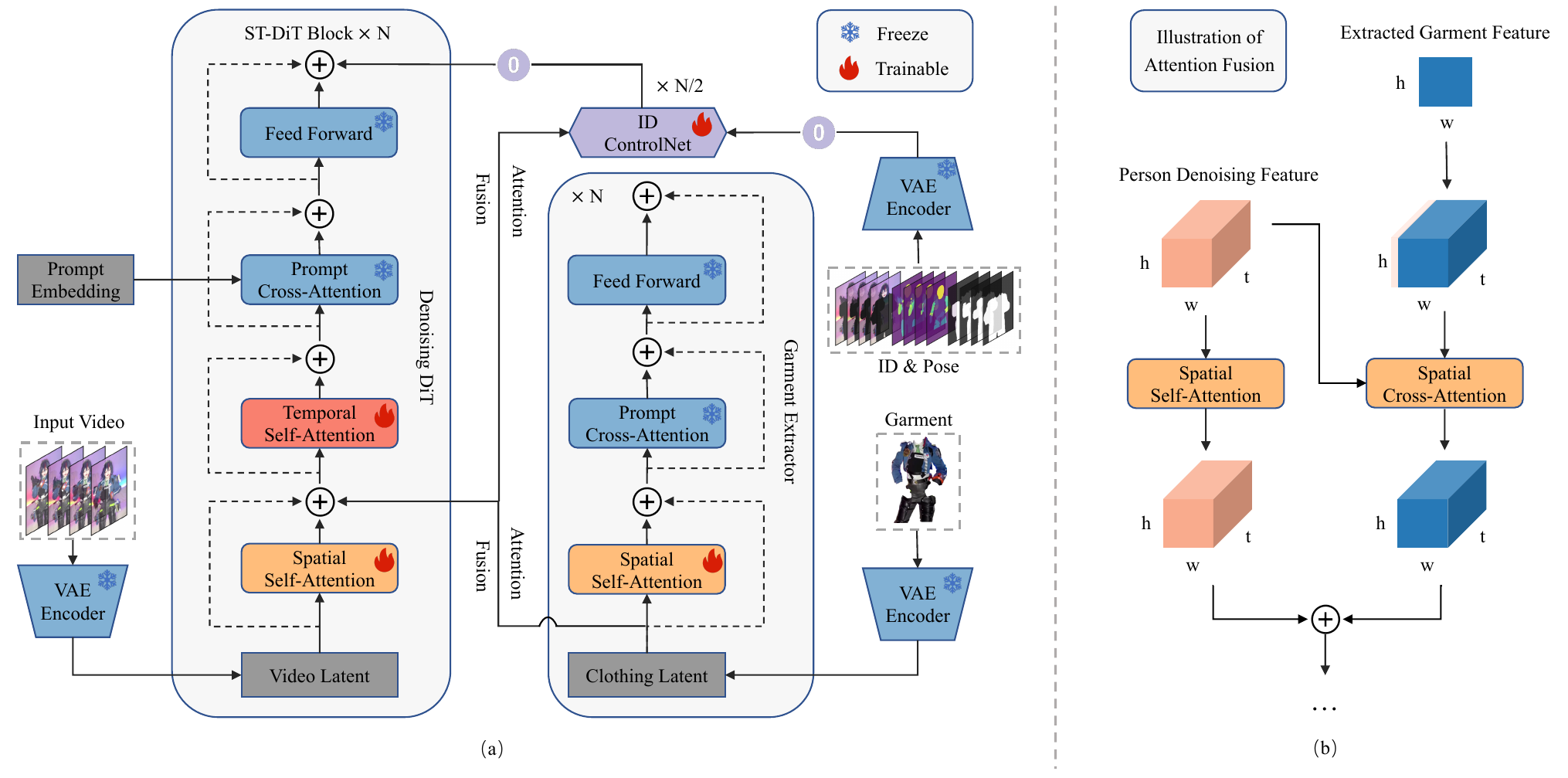}
    \vspace{-6mm}
    \caption{Overview of the proposed VITON-DiT. (a) The architecture contains three components with the following tasks. (1) \textit{Denoising DiT}: generating latent representation of video contents via a chain of Spatio-Temporal (ST-) DiT blocks. (2) \textit{ID ControlNet}: producing feature residual for the Denoising DiT to preserve the reference person's identity, pose, and background. (3) \textit{Garment Extractor}: obtaining and delivering garment features into the Denoising DiT and the ControlNet via attention fusion, thus recovering detailed clothing textures in the generated try-on video. (b) Illustrated Attention Fusion: integrating person denoising features and extracted garment features using addictive attention. This operation is utilized in both the Denoising DiT and the ID ControlNet.}
    \label{fig:framework}
\end{figure*}

We present VITON-DiT, a video virtual try-on framework built upon diffusion transformers (DiT)~\cite{Peebles2022DiT}. Our model comprises three components as shown in Fig.~\ref{fig:framework}(a): (1) the main \textbf{Denoising DiT} built by a series of \emph{Spatio-Temporal (ST-) DiT} blocks, performing latent diffusion procedure and generate video try-on results, (2) the \textbf{Garment Extractor} that processes and delivers garment features to retain clothing details, (3) the trainable \textbf{ID ControlNet }
that guides the Denoising DiT to preserve the target person's pose and identity. These three modules are connected by an operation called \emph{attention fusion}, as depicted in Fig.~\ref{fig:framework}(b). It fuses the extracted garment features with the person denoising features via summing up the attentive outputs. We find this addictive operation works well for recovering garment details in the generated videos. Before introducing our architecture, we briefly review some basic concepts of Latent Diffusion Models (LDMs) and DiT in Sec.~\ref{preliminary}. The three parts of VITON-DiT are elaborated from Sec.~\ref{ST-DiT} to Sec.~\ref{ID-ControlNet}. At Sec.~\ref{strategies}, we further elucidate a novel \emph{random agnostic condition swap} during training and an \emph{interpolated auto-regressive (IAR) technique} at inference. These strategies lead to improved long video generation up to 20 seconds.

\subsection{Preliminary}
\label{preliminary}

\textbf{Latent Diffusion Models (LDMs).} Generating high-resolution images/videos directly in the original pixel space can be computationally expensive and challenging due to the high dimensionality. Instead, LDMs~\cite{ldm} operate in a latent space where the data is represented in a more compact form. This approach leverages the power of variational autoencoders (VAEs)~\cite{vae} to encode the high-dimensional data into a latent space and then apply the diffusion process in this latent space. An image LDM typically contains three key components: (a) an Encoder $\mathcal{E}$ mapping the high-resolution image $x$ to a latent representation $z=\mathcal{E}\left(x\right)$\,, (b) a Diffusion Process involving a forward process that gradually adds noise to $z$ over $T$ time steps:
\begin{equation}
    q\left(z_t \mid z_{t-1}\right)=\mathcal{N}\left(z_t ; \sqrt{1-\beta_t} z_{t-1}, \beta_t I\right),
\end{equation} where $\beta_{t}$ is a variance schedule that controls the amount of noise added at each step; and a reverse process parameterized by a neural network (typically a U-Net~\cite{unet}) $p_{\theta}$ that learns to denoise: 
\begin{equation}
    p_\theta\left(z_{t-1} \mid z_t\right)=\mathcal{N}\left(z_{t-1} ; \mu_\theta\left(z_t, t\right), \sigma_\theta^2(t) I\right),
\end{equation} (c) a Decoder $\mathcal{D}$ maps the denoised latent representation back to the original image space: $\hat{x}=\mathcal{D}_{\left(z_0\right)}$. The training objective is typically a reconstruction loss in the latent space that minimizes the noise $\epsilon$ and the network's prediction: 
\begin{equation}
    L_{L D M}=\mathbb{E}_{z, \epsilon, t}\left[\left\|\epsilon-p_\theta\left(z_t, t\right)\right\|_2^2\right].
\end{equation}
Once Trained, we can sample $z_t$ from $p_{(z)}$ and decode it to image space with a single pass through $\mathcal{D}$.

\textbf{Diffusion Transformers (DiT).} The Diffusion Transformer~\cite{Peebles2022DiT} is an innovative architecture that leverages the strengths of diffusion models and transformers~\cite{Vaswani2017AttentionIA}. By integrating these two powerful paradigms, it aims to extend the quality, flexibility, and scalability of the traditional UNet-based LDMs \cite{ldm}. The overall formulation remains the same as the LDMs except using a transformer (instead of a UNet) to learn the denoising function $p_{\theta}$ within a diffusion-based framework. Inspired by OpenSora~\cite{opensora}, we adopt a modified Spatio-Temporal DiT (ST-DiT) as the backbone of our VITON-DiT. We find the spatio-temporal disentangled transformers are particularly well-suited for handling human video data with limited computational resources. 

\subsection{Spatio-Temporal Diffusion Transformer (ST-DiT)}
\label{ST-DiT}

As shown in Fig.~\ref{fig:framework}(a), each ST-DiT block consists of three attention layers that respectively perform Spatial Self-Attention (SSA), Temporal Self-Attention (TSA), and Prompt Cross-Attention (PCA), following by a point-wise feed-forward layer that bridges two adjacent ST-DiT blocks. The normalized input of each layer is further skip-connected with the layer's output.

Specifically, the input video $x \in \mathbb{R}^{f \times H \times W \times 3}$ is first projected into the latent space via a fixed VAE encoder $\mathcal{E}$~\cite{vqvae}, producing the video latent $z_0 \in \mathbb{R}^{f \times h \times w \times 4} = \mathcal{E}\left(x\right)$\,, where $h=H/8$, $w=W/8$, and $f$ refers to the number of frames. Given patch size $p \times p$, the spatial represented $z_0$ is then ``patchified'' into a sequence of length $s = hw / p^2$ with hidden dimension $d$, forming the input token $I \in \mathbb{R}^{f \times s \times d}$. 
Following patchify, the input video tokens $I$ are processed sequentially by SSA, TSA and TSA. In SSA, a vanilla Attention $(Q, K, V)=\operatorname{softmax}\left(\frac{Q K^T}{\sqrt{d}}\right) \cdot V$ is applied on the spatial dimension of normalized $I$ with
\begin{equation}
    Q=W_Q \cdot I_{norm}, K=W_K \cdot I_{norm}, V=W_V \cdot I_{norm}.
\end{equation}
Here, $W_Q, W_K, W_V \in \mathbb{R}^{d \times d_{\epsilon}}$ are learnable projection matrices~\cite{jaegle2021perceiver, Vaswani2017AttentionIA}. While in TSA, the self-attention is performed on the temporal dimension $f$ to perceive and integrate dynamic information of the input dance frames. Lastly, we send embedding of a default prompt ``\textit{a dancing person}'' through T5 text encoder~\cite{t5textencoder} into the PCA layer, where a cross attention is conducted between the prompt embedding and the intermediate feature from the TSA.

The denoising ST-DiT plays a vital role in refining the generated video sequences. The three attention layers in each block are tasked with texture generation/preservation, keeping temporal consistency and enhancing overall visual quality, respectively. Note during training, we load pre-trained weights from OpenSora~\cite{opensora} and freeze the PCA \& feed-forward layer to retain its original generation ability. The parameters of SSA and TSA are trainable as they need to process additional garment features or be adapted for in-the-wild human videos. The training objective for the main denoising DiT $\mathcal{P}$ is the standard latent diffusion loss:
\begin{equation}
     L_{\mathcal{P}}=\mathbb{E}_{z, \epsilon, t}\left[\left\|\epsilon-p_\theta\left(z_t, t\right)\right\|_2^2\right].
\end{equation}

\subsection{Garment Extractor}

The key to video try-on is recovering the texture details of the desired garments in the generation results. To this end, we design a garment extractor DiT $\mathcal{G}$ in parallel with the main denoising DiT, as illustrated by Fig.~\ref{fig:framework}(a). It drops the temporal attention as the input contains only a single clothing image $c$ (i.e., without temporal information). 

Similar to the denoising DiT, the clothing image is encoded by $\mathcal{E}$ and passes through $N$ garment extractor blocks. In each pass, we store the intermediate features before being sent to the SSA layer, then input them into the main DiT and the ID ControlNet. In particular, an Attention Fusion depicted by Fig.~\ref{fig:framework}(b) comes into play and associates the garment encoder with the other two modules via additive attention. 

Specifically, the extracted garment feature $r_c$ is firstly repeated along the temporal dimension, matching the shape of the person denoising feature $r_p$ from the same location in the denoising DiT and the ControlNet. Secondly, $r_p$ and $r_c$ are run through the SSA and the Spatial Cross-Attention (SCA), respectively. Lastly, the two output features are added and sent to the next blocks (TSA, PCA, etc) as detailed in Sec.~\ref{ST-DiT}. The intuition of this attention fusion is to incorporate garment features via SCA followed by a summed injection. We formulate the attention fusion process as:

\begin{equation}
    \mathcal{F}\left(r_p, r_c\right) = SSA(r_p, r_p) + SCA(r_p, r_c)
\end{equation}

A line of work~\cite{hu2023animateanyone,wang2024stablegarment,xu2024ootdiffusion,chen2024magic} has proved the effectiveness of this operation in keeping the texture details. We differ from them in our newly inserted SCA layer. By using an additional cross-attention layer instead of reusing the same SSA layer, we improve the model's capacity and capability to perceive the garment features, thereby reducing the burden of SCA and eliminating its input ambiguity.

\subsection{Identity Preservation ControlNet (ID ControlNet)}
\label{ID-ControlNet}

At its core, video virtual try-on can be viewed as an inpainting problem. It requires four-tuple $\left\{x_a, d_p, m_c, c\right\}$ to put the target clothing $c$ onto the reference person video $x$, including the cloth-agnostic image $x_a$, DensePose image $d_p$, and the inpainting mask $m_c$, as visualized in Fig.~\ref{fig:framework}(a). As the pre-trained weight from OpenSora is not tuned for inpainting, we introduce a DiT ControlNet to preserve the person's pose, identity and background, named ID ControlNet $\mathcal{C}$.

Specifically, the ID ControlNet is a trainable replica of the front half of the denoising DiT. Similar to the U-Net ContorlNet~\cite{controlnet}, two zero-initialized linear layers are plugged into the two ends of $\mathcal{C}$, as shown in Fig.~\ref{fig:framework}(a). The difference here is the denoising DiT does not have skip-connections between its internal blocks. We therefore add the outputs of $\mathcal{C}$ as residual solely to the front half of the denoising DiT, rather than the end half as in U-Net ControlNet.

Formally, given a tuple of agnoistic-conditon sequence $\left(x_a, d_p\right) \in \mathbb{R}^{f \times H \times W \times 3}$\,, the VAE Encoder $\mathcal{E}$ yields the latent $\left(z_a, z_p\right) \in \mathbb{R}^{f \times h \times w \times 4}$ that are further concatenated with the resized mask $m_c \in \mathbb{R}^{f \times h \times w \times 1}$. Then the resulting latent of size $\mathbb{R}^{f \times h \times w \times 9}$ is patchified and passed through a zero-initialized linear layer before sending to the ID ControlNet. The output signals of $\mathcal{C}$ are directly infused into the denoising DiT as feature residuals. This setup enables $\mathcal{C}$ to deliver precise, pixel-aligned control signals for accurate identity preservation. To summarize:
\begin{equation}
    r_s = \mathcal{C}\left(\mathcal{E}\left(x_a\right) \odot \mathcal{E}\left(d_p\right) \odot m_c\right), 
\end{equation}
where $\odot$ refers to concatenation. We empirically find our ID ControlNet is robust to accidental condition errors. Take Fig.~\ref{fig:data-scaling}(a) as an example, while the DensePose input of the ControlNet has obvious artifacts, our VITON-DiT still yields rational results despite inaccurate pose guidance.

\subsection{Strategies for Long Video Generation}
\label{strategies}

\begin{figure}
    \centering
    \vspace{-6mm}
    \includegraphics[width=1\linewidth]{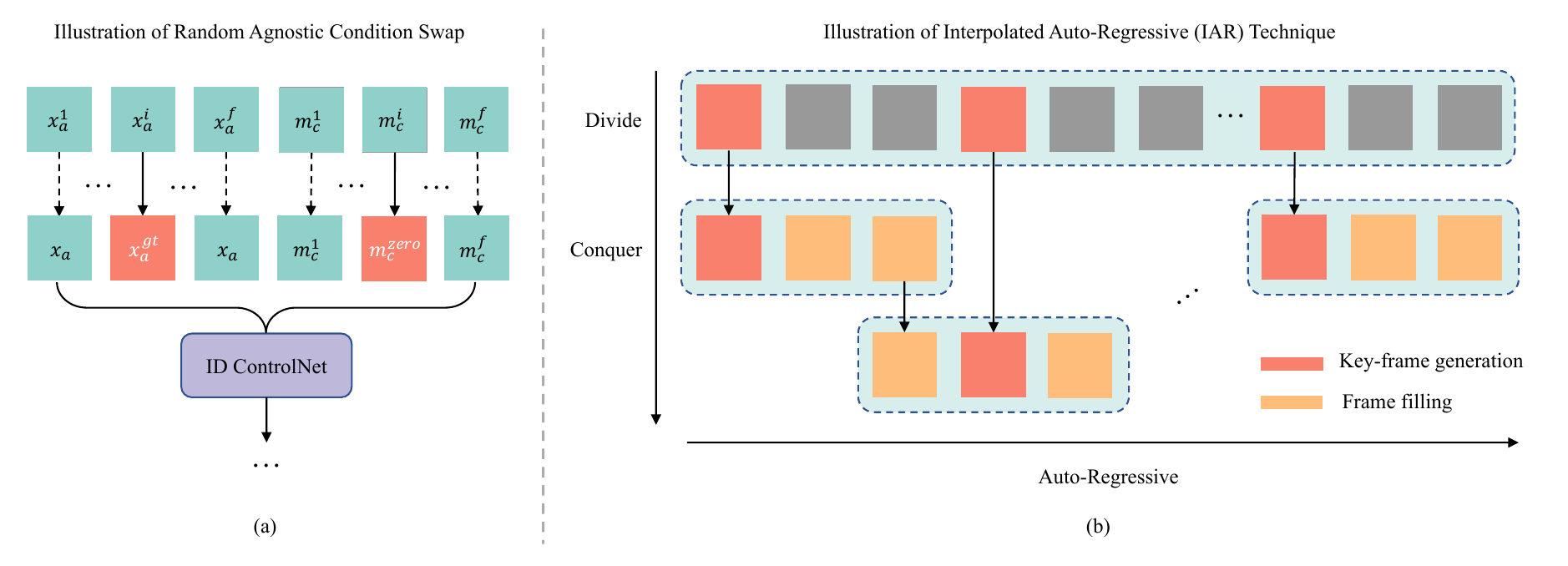}
    \caption{Strategies for long video generation. (a) Random agnostic condition swap: randomly replacing agnostic images and inpainting masks with corresponding ground-truth and all-zero masks. (b) IAR inference: generating key-frames within each divided sequence, followed by an AR inference that fills missing frames. Note that random swap training is the prerequisite for the IAR inference. } 
    \label{fig:strategy}
\end{figure}
Directly generating long videos is quite challenging especially when computational resources are limited. To mitigate this, we propose a noval random selection strategies during training and an Interpolated Auto-Regressive (IAR) technique at inference, as depicted in Fig.~\ref{fig:strategy}.

\subsubsection{Random Selection Training Strategies}
Overall, these strategies include a \emph{random agnostic condition swap} for ID ControlNet and a \emph{random stride selection strategy} for frame sampling.

\textbf{Random agnostic condition swap}: During training, we randomly sample \textit{k} frames and swap their agnostic images $x_a$ to corresponding ground-truth frames, and set their agnostic mask $m_c$ to all-zero. More precisely, this strategy is merely applied for the ID ControlNet while the input to the denoising DiT remains unchanged so as to avoid model collapse. Without the random swap, it becomes intractable to perform auto-regressive long video generation at test time. 

\textbf{Random stride selection strategy}: During training, we randomly vary the stride used when sampling frames, and the range of the selected stride is determined according to the average frame counts of clips in each respective dataset. This increases the probability of the model seeing diverse viewpoints, such as side views or back views. As a result, the model gains an improved ability to generate temporally coherent try-on videos across different body views.

\subsubsection{Interpolated Auto-Regressive (IAR) Inference.}
Long video generation typically relies on an auto-regressive (AR) approach for inference~\cite{tseng2022edge}, but this method often leads to quality degradation of the generated content over time. Inspired by the divide-and-conquer algorithm~\cite{yin2023nuwaxl}, we introduce a new IAR technique that boosts the traditional auto-regressive approach for generating high-quality extended videos. It bifurcates the video generation into two sub-tasks: key-frame generation and frame filling. 
Specifically, for generating a $f$-frame video, the IAR first divides it into $n$ sub-videos, predicts the starting frames within each sub-video based on the provided conditions, and subsequently conducts AR generation to fill the missing $f-n$ frames. The way of AR is to iteratively set the last $j$ frames of the previous video clip as the condition to generate the current clip. The key is at each iteration, we will replace the agnostic condition with the aforementioned $n$ interpolated frames if available. This prevents the model from suffering quality degradation due to occlusions, and also benefits from the AR that ensures smooth video footage, 
as evidenced in Fig.~\ref{fig:mix-ablation}.(c) and Table.~\ref{tab:abl_3}.

\section{Experiments}
\label{experiments}

\subsection{Implementation Details}


\textbf{Dataset Construction.} We collect an unpaired dataset of human dance videos that encompasses a wide variety of clothing, backgrounds and body motions. We segment the collected thousands of videos using scene detection tool \footnote{\url{https://github.com/Breakthrough/PySceneDetect}} and filer out clips with multiple persons or a small portion of people, resulting in over 15,000 high-quality video clips. We further incorporate the FashionVideo~\cite{fashionvideo} and the Tiktok dataset~\cite{Jafarian_2021_CVPR_TikTok} for training. We deliberately select 50 clips with diverse identities and backgrounds to serve as a new benchmark for evaluating video try-on outcomes.

\textbf{Multi-Stage Self-Supervised Training.}
To exploit the unpaired dataset, we progressively train the model in a self-supervised manner. We segment out the clothing in a random frame of each video clip by human parsing~\cite{cihp}, and augment it with random rotation and resizng before sending to the garment extractor. We first load pretrained weights of OpenSora, and arrange the training of the three modules in the following order: 
\begin{itemize}
    \item \emph{Image pretraining for garment extractor}: We only train the garment extractor and freeze all others to reconstruct the person image from its parsed clothing. The motivation is to expose the model to a larger volume of clothing imagery, while also reinforcing its ability to generate human images.
    \item \emph{Image pretraining for ID ControlNet}: We now incorporate the ID ControlNet and set all parameters trainable except for the SSA of the denoising ST-DiT. The training objective is same as the first stage.
    \item \emph{Video fine-tuning for VITON-DiT}: Lastly, we open all parameters for training except the SSA modules of the denoising ST-DiT.
\end{itemize}

\textbf{Hyper-parameters Setting.} We train VITON-DiT with two resolutions of 192x256 and 384x512. We use the low-resolution version for qualitative and quantitative comparison with baselines on the standard VVT dataset~\cite{fwgan} and for the ablation study. And the high-resolution model is for demo purposes. We adopt the AdamW optimizer~\cite{adamw} with a fixed learning rate of 1e-5. The models are trained on 8 A100 GPUs. In the first two training stages, we utilized unpaired image data extracted from videos, and merge them with the existing VITON-HD dataset~\cite{SeungHwanChoi2021VITONHDHV}. We set $f=36$ frames for low-resolution model and $f=16$ for the high-resolution model. We keep the same number of ST-DiT blocks $N=28$ with OpenSora, and set the patch size $p$ equal to 2, the hidden dimension $d=1152$.

\begin{figure}[h!]
    \centering
    \includegraphics[width=1\linewidth]{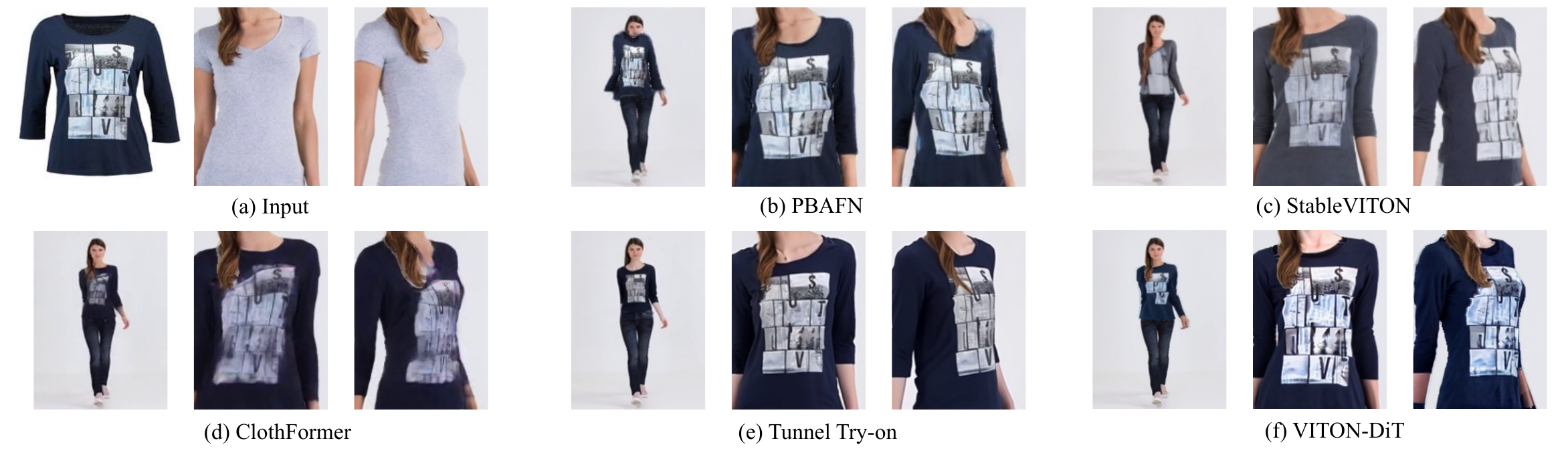}
    \vspace{-4mm}
    \caption{Qualitative comparison with baselines. Our VITON-DiT outperforms other baselines in terms of consistent preservation of garment shape and color, as well as stable clothing-person alignment cross varing camera distances.}
    \vspace{-4mm}
    \label{fig:qualitative}
\end{figure}

\subsection{Qualitative Results}
Fig.\ref{fig:qualitative} presents the visual comparison between VITON-DiT and other baselines on the VVT dataset. It is clear that GAN-based methods like PBAFN~\cite{YuyingGe2021ParserFreeVT} (Fig.~\ref{fig:qualitative}(a)) and ClothFormer~\cite{clothformer} (Fig.~\ref{fig:qualitative}(d)), are prone to clothing-person misalignment due to the inaccurate garment warping procedure. This becomes more severe for PBAFN on videos containing varied sizes of individuals. Although ClothFormer can handle smaller proportions of people, the generated images are often blurry and exhibit distorted cloth texture. The diffusion-based StableVITON~\cite{kim2023stableviton} produces relatively accurate single-frame results. However, due to the image-based training, StableVITON does not account for temporal coherence, resulting in noticeable jitters between consecutive frames (Fig.~\ref{fig:qualitative}(c)). Tunnel Try-on~\cite{xu2024tunnel} in Fig.~\ref{fig:qualitative}(e) is a concurrent work that adapts U-Net diffusion model to video try-on. Despite reasonable results, the generated clothes exhibit obvious color discrepancy compared with the input ground truth. Additionally, due to the supervised setting, it is constrained to handling product clothing images, and the provided videos on its website are only a few seconds long.

In contrast, our VITON-DiT seamlessly integrates the scalable DiT and is trained self-supervisedly, allowing for accurate single-frame try-on with high inter-frame consistency. As depicted in Fig.~\ref{fig:qualitative}(f), the letters on the chest of the clothing adhere to the input shape and color, and are correctly positioned as the subject moves closer to the camera.
Furthermore, we provide additional qualitative results using our unpaired dataset to demonstrate the robust try-on capabilities and practicality of our VITON-DiT. Fig.~\ref{fig:teaser} showcases various results generated by VITON-DiT, including scenarios involving complex dance motions against real-world backgrounds. By integrating attention fusion, our method effectively adapts to different types of human movements and clothing, resulting in high-detail preservation and temporal consistency in the generated try-on sequences. 

\begin{table}[]
    \centering
    \caption{Quantitative comparison on VVT dataset. The best results are denoted as \textbf{Bold}.}
    \begin{tabular}{ccccc}
    \toprule
    Method        & SSIM$\uparrow$   & LPIPS$\downarrow$ & $VFID_{I3D}\downarrow$ & $VFID_{ResNeXt}\downarrow$ \\ 
    \midrule
    CP-VTON~\cite{wang2018cp-vton}       & 0.459          & 0.535 & 6.361         & 12.10            \\
    FW-GAN~\cite{fwgan}        & 0.675          & 0.283 & 8.019         & 12.15            \\
    PBAFN~\cite{YuyingGe2021ParserFreeVT}         & 0.870          & 0.157 & 4.516         & 8.690            \\
    ClothFormer~\cite{clothformer}   & \textbf{0.921} & 0.081 & 3.967         & 5.048            \\
    AnyDoor~\cite{chen2023anydoor}       & 0.800          & 0.127 & 4.535         & 5.990            \\
    StableVITON~\cite{kim2023stableviton}  & 0.876          & 0.076 & 4.021         & 5.076            \\
    Tunnel Try-on~\cite{xu2024tunnel}& 0.913          & \textbf{0.054} & 3.345         & 4.614            \\
    VITON-DiT (ours)     & 0.896          & 0.080 & \textbf{2.498}        & \textbf{0.187}\footnote{Owing to the absence of open-source codes for VFID metrics, we provide our implementation at \url{https://github.com/ZhengJun-AI/vfid-metrics} for fair comparison.}              \\ 
    \bottomrule
    \end{tabular}
    \label{tab:quant}
\end{table}

\begin{table}[]
    \centering
    \caption{Ablation study for data scales and training strategies.}
    \begin{tabular}{ccccc|cc}
    \toprule
    Data-F & Data-S & TSA & SSA($\mathcal{C}$) & SSA($\mathcal{G}$) & SSIM$\uparrow$  & LPIPS $\downarrow$    \\ \midrule
    \checkmark    &          & \checkmark     & \checkmark          & \checkmark       & 0.702          & 0.259          \\
              & \checkmark        & \checkmark     &            &         & 0.681          & 0.287          \\
              & \checkmark           & \checkmark     & \checkmark          &         & 0.707          & 0.257          \\
              & \checkmark     & \checkmark     & \checkmark          & \checkmark       & \textbf{0.723} & \textbf{0.240} \\\bottomrule
    \end{tabular}
    \label{tab:abl_1}
\end{table}

\subsection{Quantitative Results}
The quantitative results are reported in Tab.~\ref{tab:quant}. For the single-frame evaluation, we adopt the Structural Similarity Index (SSIM)~\cite{2004SSIM} and the Learned Perceptual Image Patch Similarity (LPIPS)~\cite{zhang2018unreasonable} as the evaluation metrics. To assess the video-based performance, we employ the Video Fréchet Inception Distance (VFID)~\cite{fwgan} that utilizes 3D convolution to evaluate both the visual quality and temporal consistency of the generated results. We adopt two CNN feature extractors for VFID, including the I3D~\cite{i3d_metric} and the 3D-ResNeXt101~\cite{3dresnext101}.

As our training data contains a large volume of unpaired images, i.e., the clothing is parsed from the person images and may manifest defects like missing holes in Fig.~\ref{fig:teaser}, this increases the model's imagination ability yet inevitably leads to performance drop on the VVT dataset solely composed of product garment images. While the other baselines are all trained on paired dataset similar to VVT, and thus outperform the SSIM and the LPIPS image metrics. 
Despite inferior single-frame scores on the product clothing images, we beat the other methods on the more important video metrics, highlighting the advantage of our data utilization and training strategies in generating spatio-temporal consistent videos.

\begin{wraptable}{r}{6cm}
\centering
\vspace{-12mm}
\caption{Ablation study for Garment Extractor $\mathcal{G}$ and ID ControlNet $\mathcal{C}$.}
\vspace{2mm}
\begin{tabular}{cc|cc}
  \toprule
    SCA($\mathcal{P}$) & SCA($\mathcal{C}$) & SSIM$\uparrow$   & LPIPS$\downarrow$           \\ \midrule
           $\times$     &          $\times$       & 0.8520 & 0.1168          \\
           \checkmark        &    $\times$            & 0.8584 & 0.1086          \\
           \checkmark          & \checkmark          & \textbf{0.8617} & \textbf{0.1050} \\ \bottomrule
\end{tabular}
\label{tab:abl_2}
\vspace{-6mm}
\end{wraptable}
\subsection{Ablation Study}
We conduct extensive ablations for VITON-DiT on our collected dataset and the standard VITON-HD image dataset~\cite{SeungHwanChoi2021VITONHDHV} to verify the effectiveness of each components. Note that all ablation results are produced with the resolution of 192x256.

\begin{figure}
    \centering
    \includegraphics[width=1\linewidth]{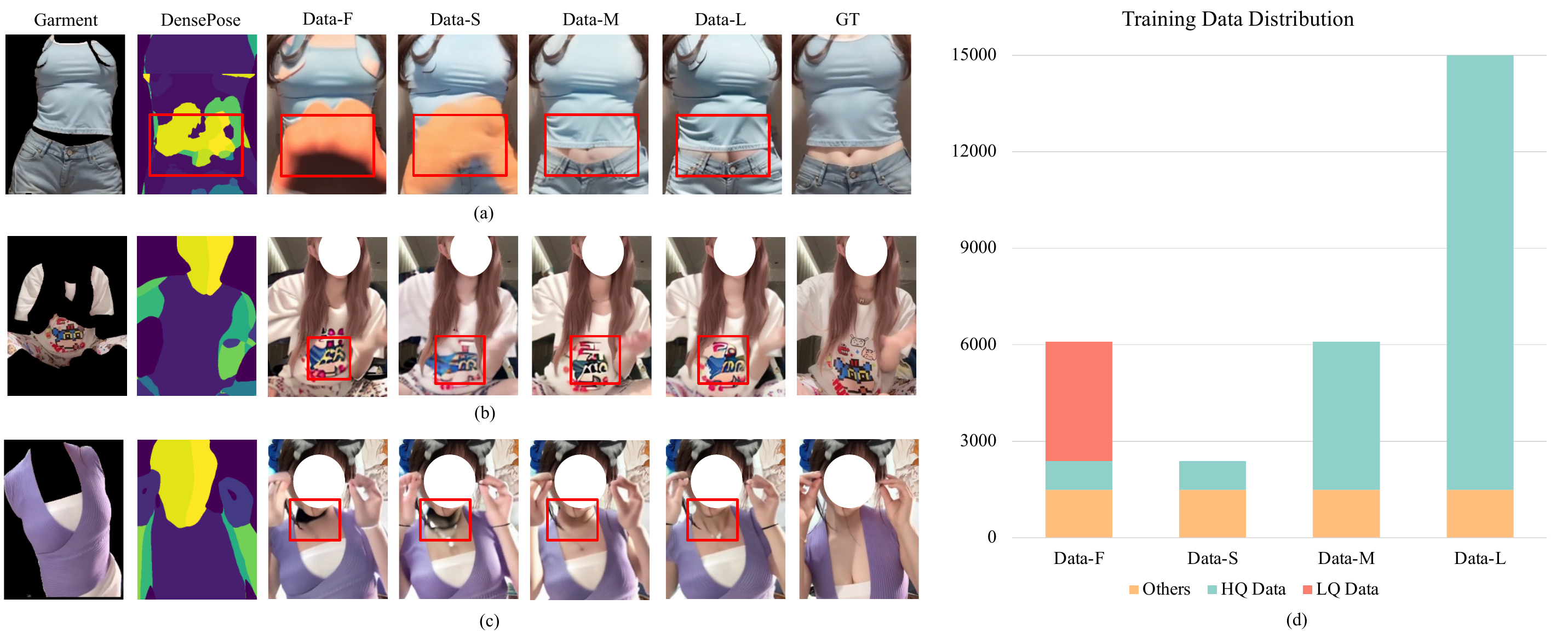}
    \vspace{-6mm}
    \caption{Ablation study on the quantity of data. It is clear that as the quality and quantity of data increases, the model's visual performance also gradually improves accordingly.}
    \label{fig:data-scaling}
\end{figure}

\subsubsection{Data Scaling}
To study the impact of data scale,
we construct four variants of training datasets from the collected videos, including the Flawed (Data-F), Small (Data-S), Medium (Data-M) and Large (Data-L) forms. As shown in Fig.~\ref{fig:data-scaling}(d), we denote the combination of VVT~\cite{fwgan}, FashionVideo~\cite{fashionvideo} and the Tiktok dataset~\cite{Jafarian_2021_CVPR_TikTok} as "Others", while refer "HQ Data" to filtered high-quality videos clips with proportion of human body exceeding $30\%$ and "LQ Data" to the low-quality unfiltered clips. 

As evident by Table.~\ref{tab:abl_1} and Fig.~\ref{fig:data-scaling}, LQ Data heavily harms the model due to the existence of extremely small or invalid persons in frames. While the model trained with even less yet high-quality data (i.e., Data-S) significantly outperforms Data-F. Moreover, a greater quantity of HQ data gains a better human body prior, allowing the model to produce reasonable results even with inaccurate pose guidance, as depicted in Fig.~\ref{fig:data-scaling} (Data-M and Data-L columns).

\begin{figure}
    \centering
    \includegraphics[width=1\linewidth]{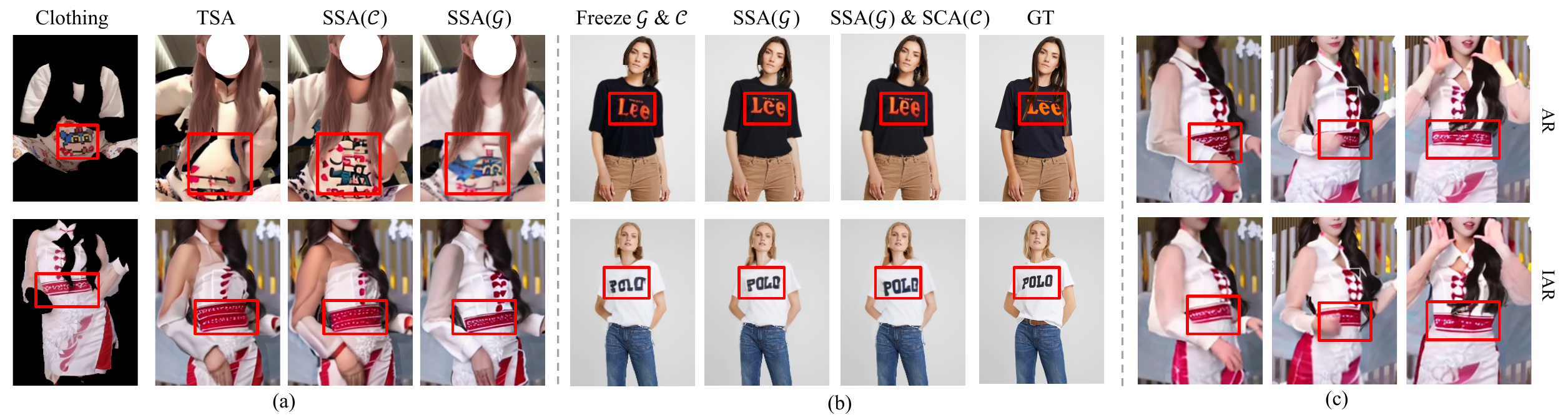}
    \vspace{-6mm}
    \caption{Ablation study on: (a) the train strategy for video fine-tuning , (b) the effectiveness of attention fusion and (c) the Interpolated Auto-Regressive (IAR) technique.}
    \label{fig:mix-ablation}
\end{figure}

\subsubsection{Training Strategy for Video Fine-tuning}
We also ablate full-tuning and freeze-tuning for the 3rd training stage. The setups are: 
\begin{enumerate}
    \item setting TSA modules of both the denoising ST-DiT $\mathcal{P}$ and the ID ControlNet $\mathcal{C}$ trainable, and freeze all other parameters;
    \item setting TSA and SSA modules of the ID ControlNet $\mathcal{C}$ trainable;
    \item setting TSA and SSA modules of both the ID ControlNet $\mathcal{C}$ and the garment extractor $\mathcal{G}$ trainable (i.e., full-tuning of $\mathcal{C}$ and $\mathcal{G}$).
\end{enumerate}
From Tab.~\ref{tab:abl_1} and Fig.~\ref{fig:mix-ablation}(a), it is clear that full-tuning is preferred to freeze-tuning, yielding more faithful results both qualitatively and quantitatively.

\subsubsection{Effectiveness of Attention Fusion}
We drop the Spatial Cross-Attention (SCA) in the ID ControlNet $\mathcal{C}$ and train the model both in freeze-tuning and full-tuning manners: (1) only train $\mathcal{C}$ and $\mathcal{G}$ while freeze SCA of the denoising DiT and (2) train all parameters except the SSA of the denoising DiT. Moreover, our full model is also added for comparison. 

Since the attention fusion mainly affects the spatial dimension, we perform this ablation right after the 2nd training stage (i.e., without TSA loaded) on the VITON-HD image dataset. As indicated in Tab.~\ref{tab:abl_2} and Fig.~\ref{fig:mix-ablation}(b), our full model with SCA in $\mathcal{C}$ is more capable of recovering the garment texuture compared to other variants.

\begin{wraptable}{r}{0.41\textwidth}
\centering
\vspace{-6mm}
\caption{Ablation study for IAR.}
\vspace{2mm}
\begin{tabular}{ccc}
  \toprule
    Method & $VFID_{I3D}$ & $VFID_{ResNeXt}$  \\ \midrule
    AR     & 3.899    & 6.486          \\
    IAR    & \textbf{3.674}    & \textbf{5.902} \\ \bottomrule
\end{tabular}
\label{tab:abl_3}
\vspace{-6mm}
\end{wraptable}

\subsubsection{Effectiveness of IAR}
Fig.~\ref{fig:mix-ablation}.(c) illustrates our proposed Interpolated Auto-Regressive (IAR) inference compared to the vanilla AR approach. Our newly proposed IAR technique enables more robust handling of occlusions and better recovery of texture details. The metrics reported in Tab.~\ref{tab:abl_3} (on our benchmark) indicate the superiority of our IAR method.

\section{Conclusions}
We propose the first DiT-based video try-on network (VITON-DiT) for facilitating scalable unpaired video virtual try-on. By utilizing an attention fusion process as well as a novel random selection and IAR strategies, VITON-DiT is able to faithfully recover clothing details in the generated videos spanning tens of seconds. Experiments highlights VITON-DiT's ability to handle diverse clothing and complex body movements, outperforming previous methods in term of spatio-temporal consistency under real-world scenarios. We believe this work will pave the way for new scalable approaches, enabling the utilization of the vast amount of readily available unlabeled video data. 

\clearpage

\bibliographystyle{plain}
\bibliography{main.bib}

\clearpage
\appendix

\end{document}